\documentclass[runningheads]{llncs}
\usepackage[T1]{fontenc}
\usepackage{booktabs}
\usepackage{algorithm}
\usepackage[switch]{lineno}
\usepackage{multirow}
\usepackage{subfig}
\usepackage{amsmath}
\usepackage{amsfonts}
\usepackage{algorithmic}
\usepackage{graphicx}
\usepackage{textcomp}
\usepackage{xcolor}
\usepackage{comment}
\usepackage{graphicx}
\usepackage{tikz}
\usetikzlibrary{arrows.meta,calc}
\usepackage{marvosym}

\begin{document}
\title{Self-Supervised Learning for Sparse Matrix Reordering}
%
%
%
%

\author{Ziwei Li\inst{1,2} \and
Tao Yuan\inst{1}\textsuperscript{(\Letter)} \and
Fangfang Liu\inst{1} \and
Shuzi Niu\inst{1}\textsuperscript{(\Letter)} \and
Huiyuan Li\inst{1}\textsuperscript{(\Letter)} \and
Wenjia Wu\inst{1}}

\institute{
Institute of Software, Chinese Academy of Sciences, Beijing, China\\
\email{\{liziwei2021,yuantao,fangfang,shuzi,huiyuan,wuwenjia\}@iscas.ac.cn}
\and
University of Chinese Academy of Sciences, Beijing, China}

\maketitle              
\begin{abstract}
Rearranging the rows or columns of a sparse matrix using an appropriate ordering can significantly reduce fill-ins, i.e., new nonzeros introduced during matrix factorization, decreasing memory usage and runtime. However, finding an ordering that minimizes fill-ins is NP-complete. Existing approaches, including graph-theoretic and deep learning methods, rely on surrogate objectives without theoretical guarantees. The Fill-Path Theorem reveals a direct and intrinsic relationship between fill-in generation and the sparse structure of the matrix as path triplet inequalities. Here we first employ a multigrid graph network to capture structural information for each vertex. We then derive a triplet sampling strategy based on inequalities. Finally, we introduce an end-max chain loss function to reduce the number of triplets whose predicted scores satisfy these inequalities. Experimental evaluations on the publicly available SuiteSparse matrix collection demonstrate the superiority of the proposed method in terms of both fill-in reduction and speedup in LU factorization time.
 
\keywords{Sparse matrix
reordering  \and Self-supervised learning \and Fill-Path Theorem.}
\end{abstract}

\section{Introduction}
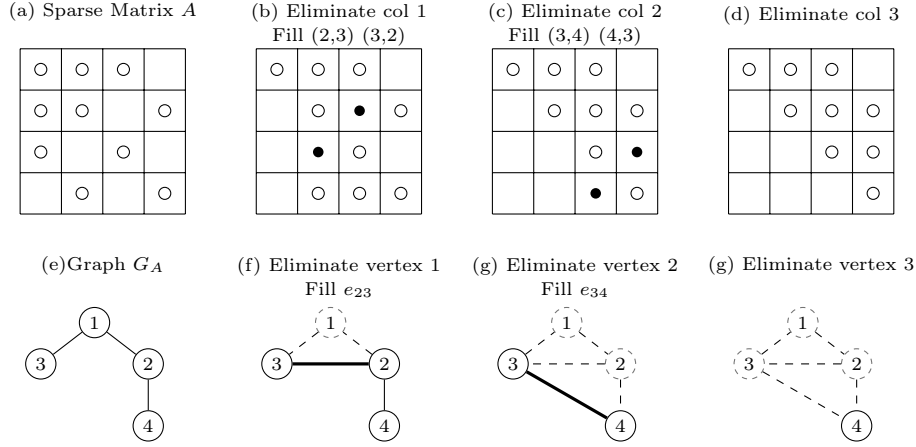
\begin{figure}[t]
  \centering
\tikzset{
  hollow/.style   ={circle,fill=black,inner sep=1.4pt},
  dot/.style={circle,draw,fill=white,inner sep=1.6pt},
  vtx/.style   ={circle,draw,minimum size=4mm,inner sep=0pt,font=\scriptsize},
  vtx-elim/.style  ={vtx, dash pattern=on 2pt off 2pt,draw=black!55, fill=white},
  vtx-pivot/.style ={vtx, very thick, fill=black!12}
}

\begin{tikzpicture}[scale=0.78]

\begin{scope}[scale=0.7]
  \node[align=center] at (2,4.6) {\scriptsize (a) Sparse Matrix $A$ \\};
  \draw (0,0) grid (4,4);
  \foreach \i in {0,1,2,3}{ \node[dot] at (\i+0.5,3.5-\i) {}; }
  \node[dot] at (1.5,3.5) {}; \node[dot] at (0.5,2.5) {};
  \node[dot] at (0.5,1.5) {}; \node[dot] at (2.5,3.5) {};
  \node[dot] at (3.5,2.5) {}; \node[dot] at (1.5,0.5) {};
\end{scope}

\begin{scope}[xshift=4.0cm,scale=0.7]
  \node[align=center] at (2,4.6) {\scriptsize (b) Eliminate col $1$ \\[-2pt] \scriptsize    Fill (2,3) (3,2)};
  \draw (0,0) grid (4,4);
  \foreach \i in {0,1,2,3}{ \node[dot] at (\i+0.5,3.5-\i) {}; }
  \node[dot] at (1.5,3.5) {}; \node[dot] at (2.5,3.5) {};
  \node[dot] at (3.5,2.5) {}; \node[dot] at (1.5,0.5) {};
  \node[dot] at (2.5,0.5) {};
  \node[hollow] at (1.5,1.5) {};
  \node[hollow] at (2.5,2.5) {};
\end{scope}

\begin{scope}[xshift=8.0cm,scale=0.7]
  \node[align=center] at (2,4.6) {\scriptsize (c) Eliminate col $2$  \\[-2pt] \scriptsize Fill (3,4) (4,3)};
  \draw (0,0) grid (4,4);
  \foreach \i in {0,1,2,3}{ \node[dot] at (\i+0.5,3.5-\i) {}; }
  \node[dot] at (1.5,3.5) {}; \node[dot] at (2.5,3.5) {};
  \node[dot] at (3.5,2.5) {};
  \node[hollow] at (2.5,0.5) {};
  \node[hollow] at (3.5,1.5) {};
  \node[dot] at (2.5,2.5) {};
\end{scope}

\begin{scope}[xshift=12cm,scale=0.7]
  \node[align=center] at (2,4.5) {\scriptsize (d) Eliminate col $3$ \\};
  \draw (0,0) grid (4,4);
  \foreach \i in {0,1,2,3}{ \node[dot] at (\i+0.5,3.5-\i) {}; }
  \node[dot] at (1.5,3.5) {}; \node[dot] at (2.5,3.5) {};
  \node[dot] at (3.5,2.5) {};
  \node[dot] at (3.5,1.5) {};
  \node[dot] at (2.5,2.5) {};
\end{scope}

\begin{scope}[yshift=-4.5cm,scale=0.7]
  \node[align=center] at (2,4.9) {\scriptsize (e)Graph $G_A$ \\};
  \coordinate (d3) at (0.5,2.8);
  \coordinate (d1) at (1.8,3.8);
  \coordinate (d2) at (3.1,2.8);
  \coordinate (d4) at (3.1,1.3);
  \node[vtx] (nd1) at (d1) {1};
  \node[vtx] (nd2) at (d2) {2};
  \node[vtx] (nd3) at (d3) {3};
  \node[vtx] (nd4) at (d4) {4};
  \draw (nd3)--(nd1)--(nd2)--(nd4);
\end{scope}

\begin{scope}[yshift=-4.5cm,xshift=4cm,scale=0.7]
  \node[align=center] at (2,4.9) {\scriptsize (f) Eliminate vertex $1$ \\[-2pt] \scriptsize Fill $e_{23}$};
  \coordinate (e3) at (0.5,2.8);
  \coordinate (e1) at (1.8,3.8);
  \coordinate (e2) at (3.1,2.8);
  \coordinate (e4) at (3.1,1.3);
  \node[vtx-elim] (ne1) at (e1) {1};
  \node[vtx] (ne2) at (e2) {2};
  \node[vtx] (ne3) at (e3) {3};
  \node[vtx] (ne4) at (e4) {4};
  \draw (ne2)--(ne4);
  \draw[dashed] (ne2)--(ne1)--(ne3);
  \draw[very thick] (ne2)--(ne3);
\end{scope}

\begin{scope}[yshift=-4.5cm,xshift=8cm,scale=0.7]
  \node[align=center] at (2,4.9) {\scriptsize (g) Eliminate vertex $2$ \\[-2pt] \scriptsize Fill $e_{34}$};
  \coordinate (f3) at (0.5,2.8);
  \coordinate (f1) at (1.8,3.8);
  \coordinate (f2) at (3.1,2.8);
  \coordinate (f4) at (3.1,1.3);
  \node[vtx-elim] (nf1) at (f1) {1};
  \node[vtx-elim] (nf2) at (f2) {2};
  \node[vtx] (nf3) at (f3) {3};
  \node[vtx] (nf4) at (f4) {4};
  \draw[dashed] (nf3)--(nf1)--(nf2)--(nf4);
  \draw[dashed] (nf2)--(nf3);
  \draw[very thick] (nf3)--(nf4);
\end{scope}
\begin{scope}[yshift=-4.5cm,xshift=12cm,scale=0.7]
  \node[align=center] at (2,4.9) {\scriptsize (g) Eliminate vertex $3$ \\};
  \coordinate (h3) at (0.5,2.8);
  \coordinate (h1) at (1.8,3.8);
  \coordinate (h2) at (3.1,2.8);
  \coordinate (h4) at (3.1,1.3);
  \node[vtx-elim] (nh1) at (h1) {1};
  \node[vtx-elim] (nh2) at (h2) {2};
  \node[vtx-elim] (nh3) at (h3) {3};
  \node[vtx] (nh4) at (h4) {4};
  \draw[dashed] (nh3)--(nh1)--(nh2)--(nh4);
  \draw[dashed] (nh2)--(nh3)--(nh4);
\end{scope}

\end{tikzpicture}
  \caption{Gaussian elimination on a symmetric sparse matrix $A$ (a–d) and the corresponding vertex elimination process on its adjacency graph $G_A$ (e–g). Black dots indicate fill-in, i.e., new nonzeros introduced during factorization, and thick edges show the corresponding fill edges added to the graph. Dashed circles and dashed edges denote vertices and edges that have already been eliminated.}
  \label{fig:fillin}
\end{figure}
Systems of sparse linear equations typically arise from discretizing and linearizing large-scale models with local interactions~\cite{saad2003iterative}. Because each unknown couples only to a small neighborhood, the assembled coefficient matrix $A$ has nonzeros in only a small fraction of entries, often $< 10\%$~\cite{Wilkinson1971}, as in computational fluid dynamics and thermal problems. The standard form is $A x = b$, where $A \in R^{n\times n}$ is sparse, $x \in R^{n}$ is the solution vector, and $b \in R^{n}$ is the right-hand-side vector.
Due to the sparsity of \( A \), the solution process primarily operates on the nonzero elements, significantly reducing storage requirements and computational complexity compared to dense matrices. For state-of-the-art sparse solvers, such as SuperLU~\cite{superLU} and MUMPS~\cite{mumps} additional steps like matrix reordering and symbolic factorization are introduced to reduce and pre-estimate the number of possible nonzeros respectively in factor matrices before de facto factorization. These nonzero entries determine the memory usage and computational cost of the subsequent sparse matrix operations.
In sparse matrix factorization, the number of nonzero entries in factorized matrices usually exceeds that of the original matrix by a factor of ten or more, substantially increasing runtime resource consumption. Fill-ins refer to new nonzero entries that appear in the factor matrices but were zero in the original matrix. LU factorization is precisely derived from the Gaussian elimination process. For each column, elementary row transformation operations are performed until the upper triangular factor matrix is obtained. Those row transformation operations are denoted as the lower triangular factor matrix. Among these operations, fill-ins are mainly generated from row subtraction operations. The number of fill-ins is computed by subtracting the nonzero count of the original matrix from that of the combined upper and lower triangular factors. The Gaussian elimination process is illustrated in Fig.~\ref{fig:fillin}(a)–(d).
On the surface, fill-ins appear to originate from numeric factorization; however, their existence is fundamentally determined by symbolic patterns and graph topology.

Rose-Tarjan's Fill-Path Theorem~\cite{Fill-Path} reveals that the emergence of fill-ins in sparse matrix factorization is a structural graph property. Each sparse matrix \(A\) is isomorphic to its adjacency graph \(G_A\). Each row/column in \(A\) corresponds to one vertex in \(G_A\), and each nonzero entry in \(A\) is mapped to an edge in \(G_A\). Eliminating one row/column in \(A\) is equal to eliminating a vertex in \(G_A\). In this sense, vertex elimination in an undirected graph \(G_A\) and Gaussian elimination on a symmetric sparse matrix \(A\) are fundamentally the same process, as shown in Fig.~\ref{fig:fillin}(e)–(g). They are represented differently but governed by identical structural rules: fill-ins propagate exclusively through connection paths and depend critically on the ordering of intermediate vertices. Under a fixed elimination ordering, a fill edge appears between two vertices if and only if there exists a path connecting them for which every intermediate vertex on that path is eliminated earlier than both endpoints. Theoretically, sparse matrix reordering is key to reducing fill-ins that originate from matrix factorization but are intrinsically determined by structural graph connectivity.

However, finding the minimal fill-in ordering is an NP-complete problem~\cite{Fill-Path,Yannakakis1981}. Within graph-theoretic reorderings, practitioners rely on local heuristics such as Reverse Cuthill–McKee (RCM) for bandwidth minimization~\cite{RCM} and Approximate Minimum Degree (AMD) for degree-based elimination~\cite{AMD}; separator-based approaches such as Nested Dissection (ND) exploit small vertex separators to expose block structure~\cite{ND}; and spectral methods~\cite{spectral} compute the Fiedler vector of the graph Laplacian to obtain a one-dimensional embedding of the vertices on the real line and then order the vertices by that coordinate.  Solving the required large sparse eigenproblem can be expensive at scale.

The number of fill-ins is too complex to be formalized explicitly and is difficult to optimize with gradient method directly. Alternatively, deep reinforcement learning methods are better at decomposing the complex objective into an iterative optimization process with the aid of environment feedback. One kind of related reinforcement learning methods, like DRL\_ND~\cite{GP}, uses graph neural network agent to act on the adjacency graph directly for graph partitioning.
This method facilitates direct leverage of the sparsity pattern with GNN agent but it was not originally proposed for fill-in minimization. 
The other kind, like AlphaElim~\cite{alphaelim}, treats the sparse matrix with a convolutional network agent and trains the network parameters by minimizing locally computed fill-ins from Gaussian elimination. However, it ignores global sparsity structure and wastes substantial computational resources on redundant numerical operations.
While both lines of work can reduce fill-in, they are RL-based and thus require many decision steps per instance, leading to long inference latency. In contrast, UDNO~\cite{UDNO} uses a GNN to directly predict a vertex permutation in a single shot, substantially cutting runtime. Its objective penalizes rank differences between adjacent vertices—an implicit bandwidth surrogate that shrinks the set of potential fill locations—but it still does not optimize fill-in according to its generative mechanism.

To avoid the inference latency of reinforcement learning and to act directly on the mechanism that creates fill-in, we propose a self-supervised reordering method, referred to as CFP, that is explicitly designed to be consistent with the Fill-Path Theorem. 
CFP first converts the sparse matrix into its adjacency graph and feeds it into multigrid graph neural networks. These networks operate on coarsening/refinement hierarchies to capture long-range structural information on large graphs. CFP is trained without ground-truth permutations; instead, it derives supervision directly from the Fill-Path Theorem, which characterizes exactly when a new fill edge is introduced. We sample triplets $(i,k,j)$ in which $k$ lies on a path between two nonadjacent vertices $i$ and $j$, and impose an end-max chain loss that encourages each such path to contain an interior vertex that is not eliminated earlier than at least one endpoint. 
This single-shot approach exploits global structure, targets fill generation at its source, and avoids multi-step RL rollouts. Experiments show that CFP achieves lower fill-in ratios and higher factorization time speedups than both graph-theoretic orderings and learning-based baselines, improving the efficiency of sparse direct solvers.

\section{Related Work}
\label{related work}
Early matrix reordering algorithms usually focus on graph-theoretic methods, such as Approximate Minimum Degree~\cite{AMD} and Nested Dissection~\cite{ND}. The rapid advancement of deep learning has driven growing interest in deep learning-based matrix reordering techniques. Here we briefly introduce these two major kinds of reordering methods.

\subsection{Graph-Theoretic Methods}
From the graph perspective, the matrix reordering problem is NP-complete for fill-in minimization. Many graph-theoretic methods have been proposed to design alternative optimization objectives. 
Next, we review three families of graph-theoretic reorderings—bandwidth-oriented schemes, degree-driven heuristics, and separator/graph-partitioning methods—and also discuss a representative spectral ordering that sorts vertices by a Laplacian Fiedler embedding.

Cuthill-McKee (CM)~\cite{CM} algorithm and Reverse Cuthill-McKee (RCM)~\cite{RCM} are commonly used graph-theoretic methods to reduce matrix bandwidth. Since fill-ins can only occur within the bandwidth, reducing the bandwidth by placing adjacent vertices close together in the ordering helps to reduce fill-ins. CM first chooses a pseudo-peripheral vertex, performs a breadth-first search of the adjacency graph \(G_A\) and sorts vertices within each level in ascending degree order. RCM attempts to reverse the whole ordering.

Minimum Degree algorithm (MD)-like methods are major graph-theoretic reordering methods to reduce fill-ins, such as MD~\cite{rose1972graph}, MMD~\cite{MMD}, AMD~\cite{AMD} and COLAMD~\cite{davis2004colamd}. MD iteratively chooses the vertex with the lowest degree in the reduced graph. This is because the fewer the neighbors of the eliminated vertex, the smaller the clique formed during its elimination, and thus fewer fill-ins are introduced. MMD algorithm incorporates more sophisticated vertex selection strategies into MD approach. AMD offers a more efficient approximation over MMD. As an extension of AMD, the COLAMD algorithm focuses on minimizing fill-ins for sparse matrix factorization by considering both row and column permutations. 

Graph partitioning methods aim to reduce fill-ins by minimizing the number of connections between subgraphs, thereby eliminating most of the potential fill-ins caused by interactions across subgraphs. The Multilevel Graph Partitioning approach is such a hierarchical method that coarsens the graph into a smaller one, partitions it, and then incrementally uncoarsens and refines this partition~\cite{karypis1998fast}. Nested Dissection (ND)~\cite{ND} involves dividing the graph into two roughly equal parts by removing a small set of vertices, known as separators, which are chosen to minimize the normalized vertex separator. This division process is recursively applied to two resultant subgraphs. The final ordering is concatenation of the vertex orderings in left, right subgraphs and the separator graph. Software packages like METIS~\cite{METIS} and Scotch~\cite{scotch} implement ND method with additional techniques to enhance the overall quality of the partitioning and ordering. Graph partitioning techniques not only help reduce fill-ins but also facilitate parallel reordering of independent subgraphs, significantly improving efficiency for large-scale sparse matrices.

Spectral ordering computes a one-dimensional embedding from the Fiedler vector of the Laplacian $L$ built on the adjacency graph \(G_A\), whose vertices index rows/columns and whose edges indicate structural nonzeros. The Fiedler vector is the minimizer of the Rayleigh quotient $\min_{u\perp \mathbf 1,\ \|u\|_2=1} u^\top L u$. Sorting vertices by the entries of this vector places adjacent vertices close in the order, typically compressing bandwidth and profile and thus reducing fill-in. It is effective on mesh-like or diffusion-dominated problems and can be combined with local post-processing. Its dominant cost---solving a large sparse eigenproblem---is computationally expensive at scale, and the method offers no provable optimality or approximation guarantees for fill-in.

\subsection{Deep Learning Methods}
Deep learning techniques have been widely used to help solve sparse linear systems~\cite{NeuralSparsel,SurveyIterative}. Among them, some studies have focused on fill-in optimization-related tasks. Deep neural networks are usually employed to estimate the possible number of fill-ins from the LU factorization of a given matrix. Combined with reinforcement learning process, deep reinforcement learning methods are used directly for matrix reordering task. More recently, deep graph neural networks reduce fill-in by optimizing surrogate objectives that suppress the set of potential nonzero locations in the factorization.

Due to the complexity of the fill-in minimization, deep neural networks are directly used to predict the number of fill-ins generated from LU factorization given a matrix for simplicity. Booth and Bolet~\cite{Acceleration} designed a neural network to select those graph-theoretic reordering methods mentioned above. Their method employs graph neural networks to learn a set of parameters for each traditional graph-theoretic reordering algorithm, with the goal of predicting the number of nonzeros introduced during matrix factorization under different reorderings. The strategy that yields the fewest fill-ins is then selected. However, this approach does not introduce a new reordering method. 

Deep reinforcement learning is to find an effective strategy in a large search space due to the combination of function approximation and strong representation learning of deep learning methods. Two typical methods are proposed to solve the matrix reordering task, including the graph-based and matrix-based deep reinforcement learning methods.
DRL\_ND~\cite{GP} treats the sparse matrix as a graph and utilizes a graph neural network agent to minimize the normalized cut or the vertex separator in a unified reinforcement learning framework. 
However, the relationship between the normalized cut (or the vertex separator) and the number of fill-ins remains unknown.
AlphaElim~\cite{alphaelim} takes matrices as images and uses a convolutional neural network-based agent, taking into account both the numeric values and the sparsity structure. The agent performs the row and column elimination process in the Monte Carlo Tree Search-based reinforcement learning framework for fill-in reduction. Numeric values, almost useless for fill-in reduction, bring about more memory consumption.

A recent deep GNN approach, UDNO~\cite{UDNO}, injects stochasticity into the vertex-scoring network to induce a distribution over permutations. It trains by minimizing the expected edge-wise rank distance, a surrogate that encourages adjacent vertices to be placed near each other and thus shrinks the set of potential fill locations. While intuitive and effective in practice, this surrogate yields only an approximate minimization: UDNO does not directly optimize fill-in or address its generative mechanism and provides no theoretical guarantees of fill-in minimization.

\section{Fill-in Generation}
Fill-in consists of the structural nonzeros that appear in the factors but not in the original matrix, i.e., the difference in sparsity pattern between $A$ and its LU factors. Our empirical and theoretical analyses indicate that the root cause of fill-in is governed by graph connectivity together with the elimination order, rather than by the specific numerical values.

\begin{figure}[htbp]
    \centering
    \subfloat[L]{
    \includegraphics[width=0.25\textwidth]{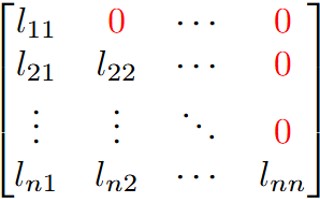}
    }
    \subfloat[U]{
\includegraphics[width=0.27\textwidth]{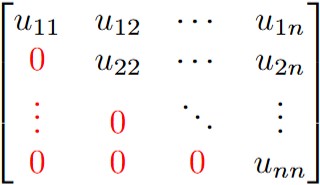}
    }    
        \subfloat[L+U-I]{
    \includegraphics[width=0.275\textwidth]{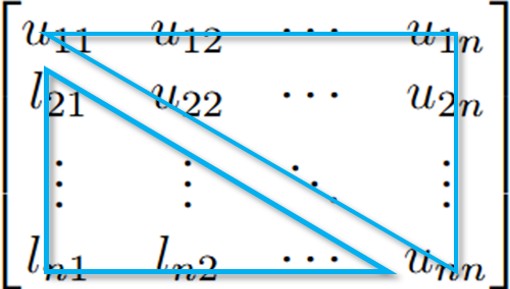}
    }
    \caption{The Combined Nonzero Pattern of the LU Factors}
    \label{fig:L+U}
\end{figure}

Given a sparse matrix \(A\), \emph{fill-in} refers to the new nonzeros introduced during factorization, e.g., LU factorization~\cite{Duff1986Direct}. In the nonsymmetric case we write \(A=LU\), where \(L\) is unit lower triangular and \(U\) is upper triangular, obtained via Gaussian elimination. To reduce storage, \(L\) and \(U\) can be kept in a single matrix by overlaying their strictly lower/upper parts—conceptually \(L+U-I\) (see Fig.~\ref{fig:L+U}). For simplicity, we mainly consider the symmetric positive definite case \(A\in\mathbb{R}^{n\times n}\). In this setting the Cholesky factorization \(A=LL^{\top}\) applies, with \(L\) lower triangular computed as Alg.~\ref{alg:cf}; the corresponding upper factor is \(U=L^{\top}\). Cholesky is the specialization of Gaussian elimination to symmetric positive definite matrices; for brevity, we will still refer to it as LU factorization.

According to Alg.~\ref{alg:cf}, the basic idea is to transform $A$ to an upper triangular matrix $L^T$ by sequentially eliminating nonzero entries below the diagonal. All entries for each column below the diagonal are subtracted from subsequent rows, leading to zeros. The number of nonzero entries in \(A\) is denoted as \(\text{nnz}(A)\). 
Fill-ins are defined as new nonzero elements that are in \(L+U-I\) but not in \(A\), and those nonzero comparison results are depicted in Fig.~\ref{fig:L+U}. 
In the context of the Cholesky factorization, subfigure (c) depicts \(L+U\) with its common diagonal removed, since \(L\) and $U=L^T$ share identical diagonal entries.
There are basically three operations in Alg.~\ref{alg:cf}: a square root computation in Line 3, a division by \(L_{kk}\) in Line 5 and the actual row subtraction in Line 7. The first two operations clearly do not introduce new nonzeros nor do they destroy them, so the inherent fill-in reason lies in the row subtraction, i.e., the relationship between rows.
\begin{algorithm}[!htbp]
\caption{Numerical Cholesky Factorization Algorithm}
\label{alg:cf}
   \begin{algorithmic}[1]
      \REQUIRE Symmetric positive definite matrix $A\in\mathbb{R}^{n\times n}$
      \ENSURE Lower triangular matrix $L$
      \STATE $L = A$
      \FOR{$k=0$ \textbf{to} $n-1$}
      \STATE $L_{kk}\gets\sqrt {L_{kk}}$
      \FOR{$i=k+1$ \textbf{to} $n-1$}
      \STATE $L_{ki}\gets L_{ki}/L_{kk}$
      \FOR{$j=i$ \textbf{to} $n-1$}
      \STATE $L_{ij} \gets L_{ij} - L_{ki}L_{kj}$
      \ENDFOR
      \ENDFOR
      \ENDFOR
      \STATE $ L = triu(L)^T$
   \end{algorithmic}
\end{algorithm}

In fact, fill-ins are implicitly embedded in sparsity structures, described in the adjacency graph $G_A$ of matrix $A$ as Def.\ref{def:graph}. When \(A\) is symmetric, \(G_A\) is an undirected graph. In this paper, we focus on the symmetric matrix \(A\). Each row or column of \(A\) corresponds to a vertex in \(G_A\). The number of rows or columns in \(A\) is equal to the vertex number of \(G_A\). Each nonzero entry $A_{ij}\neq 0$ is an edge $e_{ij}\in E_A$ in $G_A$.  
\begin{definition}
\label{def:graph}
An \(n\times n\) matrix $A$ induces the adjacency graph $G_A=(V_A,E_A)$, where 
\[
V_A=\{0,1,\ldots,n-1\},
\]
\[
E_A=\{(i,j)|0\leq i,j\leq n-1,A_{ij}\neq 0\}.
\]
\end{definition}

Fill-Path Theorem~\cite{Fill-Path} establishes a mathematical relationship between the nonzero structure of the original matrix and the factored matrix with the aid of the adjacency graph, forming the theoretical foundation of symbolic factorization~\cite{markowitz1957elimination,rose1972graph}. 
It states that each entry \(L_{ij}\) of the factor matrix is estimated as nonzero if and only if there is a path between vertices \(i\) and \(j\) in \(G_A\), such that all vertices in the path have elimination order indices smaller than both \(i\) and \(j\). The precise statement is described in Theorem~\ref{thm:fill-path}. It's worth noting that the natural ordering of \(A\) is used as elimination ordering for simplicity in this statement. Thus the equivalence among fill-ins, filled edges \(\mathcal{F}\) and multi-hop paths \(\mathcal{P}_{ij}\) is clear as Eq.(\ref{eq:3}).
\begin{theorem}[Fill-Path Theorem (Rose \& Tarjan, 1975)]
\label{thm:fill-path}
Let $A$ be a symmetric sparse matrix with associated undirected graph $G_A = (V_A, E_A)$. 
During \textsc{Cholesky} factorization with elimination order $1 < 2 < \cdots < n$, 
a \textbf{fill-in} occurs at position $(i,j)$ ($i < j$) in the factor matrix 
\textbf{if and only if} there exists a path 
\begin{equation}
i = v_0 \to v_1 \to \cdots \to v_k = j 
\end{equation} 
in \(G_A\) such that all intermediate vertices satisfy 
\begin{equation}
v_m < \min(i,j) \quad \forall m = 1, \dots, k-1.
\end{equation}
\end{theorem}
\begin{equation}
\begin{aligned}[c]
L_{ij}\neq 0\Leftrightarrow\mathcal{F}\cup\{(i,j)\}\Leftrightarrow 
\mathcal{P}_{ij}=(i , v_1,\ldots,v_{k-1}, j)\\
v_m<min(i,j),0<m<k
\end{aligned}
\label{eq:3}
\end{equation}

As mentioned before, row or column relations are key to cancellation in numeric factorization process like Alg.~\ref{alg:cf}. These relations are intuitively described as paths in its adjacency graph. The connective path between vertices \(i\) and \(j\) in Eq.~(\ref{eq:3}) visually models possible row/column operations of element \(L_{ij}\). So the intuitive understanding of Theorem~\ref{thm:fill-path} is that if path endpoints \(i\) and \(j\) are eliminated later than all these intermediate vertices, there must exist some row/column operation making \(L_{ij}\) a new nonzero and a direct edge \((i,j)\) must be added. 

The simplest case is that there is a path \(i\rightarrow k\rightarrow j\) before the \(k<\text{min}(i,j)\)-th vertex in the elimination sequence is removed. This path is graphically described the equation in Line 7 of Alg.~\ref{alg:cf} as follows in Eq.~(\ref{eq:cf2gcf}). The original \(L_{ij}\) on the right side of the equation is \(A_{ij}=0\). The joint path of \(i\rightarrow k\) and \(k\rightarrow j\) is formalized as \(L_{ik}L_{jk}\). Removing the \(k\)-th vertex occurs concurrently with both edges \((i,k)\) and \((j,k)\) deletion, which is expressed as \(L_{ij}-L_{ik}L_{jk}\) leading to a nonzero value. Thus a fill-edge is added between the \(i\)-th and \(j\)-th vertex. A specific example is shown in Fig.\ref{fig:fillin} aligning the row/column subtraction with the vertex elimination in its adjacency graph. The cancellation of row \(1\) is equal to the elimination of vertex 1, generating edge \((2,3)\) and $L_{23}=U_{32} \neq 0$ due to the path \(2\rightarrow 1\rightarrow 3\).
\begin{equation}
L_{ij}=L_{ij}-L_{ik}L_{jk}\quad \sim\quad  i\rightarrow k\rightarrow j
    \label{eq:cf2gcf}
\end{equation}

For each vertex in the elimination sequence, we construct such a simple path from its neighbors in the adjacency graph and precompute whether the corresponding entry is zero based on the theorem. The method enumerates all fill-ins by tracking all connection paths given an elimination ordering.

\section{Method}

We propose a self-supervised matrix reordering method that optimizes fill-in reduction by leveraging the relationship between fill-in generation and the sparse structure of the matrix, as characterized by the Fill-Path Theorem. Specifically, the sparse matrix \(A\) is first transformed into its corresponding adjacency graph \(G_A\), which is then fed into multigrid graph neural networks (GNNs) to capture structural information and produce vertex scores. Subsequently, a triplet sampling strategy is designed to randomly select constrained vertex triplets from the graph, and based on the fill-in generation conditions described in the Fill-Path Theorem, we construct an end-max chain loss function to guide the optimization.

\subsection{Network Architecture}
Our method uses two multigrid graph neural networks~\cite{Spectral Embedding,UDNO} to obtain vertex scores, which are then used to produce the final reordering of the sparse matrix. Leveraging a coarsen-then-refine hierarchy with graph convolution layers, the multigrid GNN effectively captures long-range, global information on large graphs that is otherwise difficult to model. The overall architecture comprises a Spectral Embedding module and a Vertex Encoder module. 

We first convert the sparse matrix $A$ into its adjacency graph $G_A$, which serves as the input. Starting from $G_A$, we repeatedly coarsen the graph using Graclus-style clustering until only two vertices remain. These two coarse vertices are initialized with the feature vectors $[1,0]$ and $[0,1]$. We then interpolate these features back to the original graph size by progressively uncoarsening the hierarchy. During interpolation, we apply SAGEConv~\cite{sageconv} layers to aggregate vertex features. After interpolation, a linear layer followed by a QR decomposition produces the output. We denote this mapping as
\(
x \;=\; S_{e}(G_A),
\)
where $S_{e}$ is the Spectral Embedding network with parameters $e$. This module is trained to approximate the Fiedler vector, i.e., the eigenvector associated with the smallest nonzero eigenvalue of the graph Laplacian, thereby capturing global structural information. The resulting approximation $x$ is used as vertex features for the Vertex Encoder module.

The Vertex Encoder module uses the same type of multigrid-like GNN architecture. Unlike the Spectral Embedding module, SAGEConv is applied not only during interpolation (uncoarsening) but also in the coarsening process. After interpolating back to the original graph, an additional SAGEConv layer and a linear layer directly output the final vertex score predictions. We write this as
\(
y \;=\; f_{\theta}(x),
\)
where $f_{\theta}$ is the Vertex Encoder network with parameters $\theta$, and $y$ denotes the per-vertex scores used to define the elimination ordering.

\subsection{Triplet Sampling and End-Max Chain Loss}

According to Theorem~\ref{thm:fill-path}, we use the natural row/column ordering of the matrix as the elimination order. In this view, the vertex index also indicates its elimination step: a larger index means the vertex is eliminated later.
The theorem states that a fill-in between two vertices $i$ and $j$ appears if and only if there exists a path from $i$ to $j$ in $G_A$ whose intermediate vertices are all eliminated earlier than both $i$ and $j$. Equivalently, if $i$ and $j$ are not directly connected and we want to avoid creating a new fill edge $(i,j)$, then for every path between $i$ and $j$ there must be at least one internal vertex $k$ on that path that is eliminated after at least one of the two endpoints. Consequently, our triplet sampling strategy randomly draws vertex triplets $(i,k,j)$ from $G_A$ and retains those that satisfy: (1) $i$ and $j$ are not adjacent, i.e., $(i,j)\notin E_A$, yet there exists at least one path between them; and (2) $k$ is an internal vertex on an $i$–$j$ path. 
\begin{figure}
    \centering
    \subfloat[Sampling a Path]{
    \includegraphics[width=0.25\textwidth]{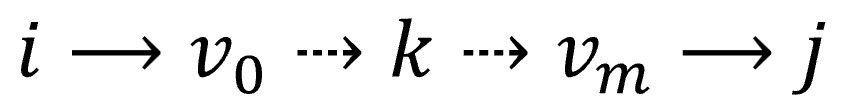}
    }
        \subfloat[Sampling a triplet]{
    \includegraphics[width=0.3\textwidth]{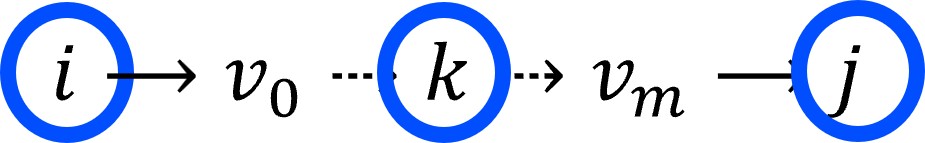}
    }
        \subfloat[End-max Chain Constraint]{
    \includegraphics[width=0.35\textwidth]{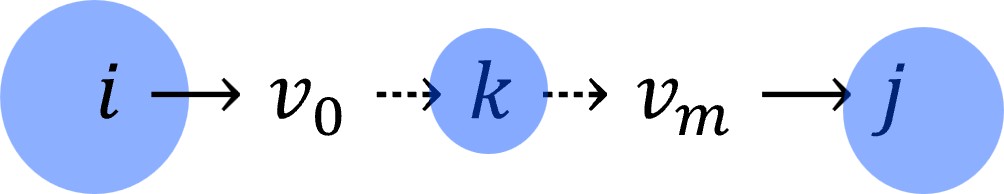}
    }
    \caption{Illustration of (a)sampling a path from the graph; (b) sampling a triplet of circled entries from the path; (c) adding end-max chain constraint by pushing the end vertex towards large prediction score with a big shadow circle.}
    \label{fig:chain}
\end{figure}

Like Fig.~\ref{fig:chain}(c), we then construct an end-max chain loss by imposing constraints on these triplets so as to discourage the creation of the fill edge $(i,j)$ in accordance with Theorem~\ref{thm:fill-path}.
Our network assigns each vertex $u$ a scalar score $y_u$. A higher score means that $u$ should be eliminated earlier (i.e., it has higher elimination priority), and a lower score means it should be eliminated later. 
For each sampled triplet $(i,k,j)$ that satisfies the path-based eligibility conditions, we would like $k$ to be eliminated after at least one of $i$ or $j$. Since a lower score corresponds to later elimination, this requirement can be enforced by encouraging
\(
    y_k \;<\; \max \bigl( y_i,\, y_j \bigr).
\)

In practice, we implement this preference as a pairwise ranking objective using a Bradley--Terry model together with a binary cross-entropy loss with logits (BCEWithLogitsLoss).
Formally, for each sampled triplet $(i,k,j)$, we define the end-max margin as
\[
    m_{i,k,j} = \max(y_i,\, y_j) - y_k,
\]
which quantifies how much later vertex $k$ is predicted to be eliminated than at least one of the endpoints $i$ or $j$. A positive margin indicates that $k$ satisfies the desired elimination order, while a negative margin suggests that $k$ is eliminated too early, potentially introducing a new fill edge $(i,j)$.
Following the Bradley--Terry model, we interpret this margin as the log-odds of a probabilistic comparison:
\[
    P(y_k \;<\; \max \bigl( y_i,\, y_j \bigr))
    = \sigma(m_{i,k,j})
    = \frac{1}{1 + e^{-m_{i,k,j}}},
\]
where $\sigma(\cdot)$ denotes the sigmoid function. To encourage $m_{i,k,j} > 0$, we minimize the negative log-likelihood under this model, leading to the following binary cross-entropy objective with logits:
\[
    \mathrm{BCEWithLogitsLoss}(m_{i,k,j}, 1)
    = \max(m_{i,k,j}, 0) - m_{i,k,j} + \log(1 + e^{-|m_{i,k,j}|}).
\]
The total end-max chain loss is obtained by averaging over all sampled triplets:
\[
    \mathcal{L}_{\text{end-max}}
    = \frac{1}{N_{\text{samples}}} \sum_{(i,k,j)} 
    \mathrm{BCEWithLogitsLoss}(m_{i,k,j}, 1).
\]

This formulation encourages the intermediate vertex $k$ to receive a smaller score than at least one endpoint, thereby aligning the learned elimination order with the sufficient condition of Theorem~\ref{thm:fill-path} and reducing the likelihood of generating new fill-ins. Because these constraints are derived directly from the Fill-Path Theorem and require no ground-truth permutations or labeled elimination orders, the resulting training objective is self-supervised. By enforcing this constraint over sampled $(i,k,j)$ triplets across the graph, the end-max chain loss biases the model toward fill-path–consistent elimination orders. The overall training procedure is illustrated in Alg.~\ref{alg:training}.
\begin{algorithm}[t]
\caption{Training Multigrid Graph Networks with End-max Chain Loss}
\label{alg:training}
\begin{algorithmic}[1]
\REQUIRE Sparse symmetric matrix $A$, Stage~I network parameter $e$
\ENSURE Stage~II network parameter $\theta$
\STATE Build the adjacency graph $G_A$ from $A$
\STATE Run the two-stage MgGNN on $G_A$ to obtain scores $y=f_{\theta}(S_e(G_A))$
\STATE $\mathcal{L}_{\text{end-max}} \gets 0$
\FOR{$t = 1$ to $T$}
  \STATE Sample distinct vertices $(i,k,j)$ from $V_A$
  \IF{$(i,j)\notin E_A$ \textbf{and} $k$ is an internal vertex on some $i$--$j$ path in $G_A$}
    \STATE $m_{i,k,j} \gets \max\!\bigl( y_i, y_j\bigr) - y_k$
    \STATE $\mathcal{L}_{\text{end-max}} \gets \mathcal{L}_{\text{end-max}} + \mathrm{BCEWithLogitsLoss}(m_{i,k,j}, 1)$
  \ENDIF
\ENDFOR
\STATE Update $\theta$ by backpropagating $\nabla_\theta \mathcal{L}_{\text{end-max}}$(Adam)
\end{algorithmic}
\end{algorithm}

\section{Experiments}

We compare the proposed method with representative traditional graph-theoretic methods and learning-based methods in terms of fill-in and speedup in LU factorization time, demonstrating the effectiveness of mitigating fill-in from its generative mechanism. We also conduct ablation studies to validate the necessity and contributions of each architectural component.
\subsection{Experimental Details}
Following prior work~\cite{Spectral Embedding,UDNO}, our training set comprises three categories: (1) matrices obtained from triangulations on various geometric domains; (2) finite-element (FEM) matrices; and (3) 2D/3D problem matrices from the SuiteSparse Matrix Collection~\cite{suitesparse}. The training matrices range from \(100\) to \(5000\) in size. For testing, we randomly sample more than 100 large-scale matrices (dimension \(>10{,}000\)) from SuiteSparse, covering Structural Problems (SP), Computational Fluid Dynamics (CFD), Model Reduction (MRP), 2D/3D discretizations (2D3D), and Thermal Problems (TP).

All experiments are conducted on an NVIDIA GeForce RTX 4090 GPU. We perform a logarithmic grid search for the learning rate in \([10^{-5}, 10^{-1}]\) and select \(10^{-5}\) as the final choice. All the network backbones with hidden dimension sizes are \(16\). The sampled triplet size is set to \(10\) times of the row or column number. Detailed information will be in the public released code\url{}.

We consider the following baselines. Graph-theoretic methods: AMD~\cite{AMD}; METIS(ND)~\cite{METIS}, i.e., Nested Dissection as implemented in METIS; and Fiedler~\cite{spectral}, a spectral reordering based on the Fiedler vector. Learning-based methods: \(S_e\), which uses the Stage~I approximation of the Fiedler vector directly for reordering; GPCE, which takes the output of \(S_e\) as vertex features, applies two SAGEConv layers, and is supervised using the best (minimum-fill) ordering among classical methods as a proxy ground truth via pairwise cross-entropy loss; and UDNO~\cite{UDNO}, an architecture with a network similar to ours but optimized to shrink post-reordering vertex distances.

We report two metrics. The first is the fill-in ratio (FIR), defined as follows.
\(L\) and \(U\) are the LU factors, and \(I\) is the identity. This quantity normalizes fill-in across matrices of different sizes and sparsity patterns. The factorization speedup is computed as the ratio between the factorization times of the original and the reordered matrices. Natural LU factorization time $t_{\text{naturalLU}}$ is the wall-clock time for LU factorization of the original matrix $A$ without reordering. $t_{\text{LU}}$ and $t_{\text{reorder}}$ are the wall-clock times required to factorize and obtain the reordered matrix, respectively. They reflect the computational cost of the downstream direct solve.
\[
\mathrm{FIR} = \frac{\mathrm{nnz}(L+U-I) - \mathrm{nnz}(A)}{\mathrm{nnz}(A)} ,
\mathrm{speedup} =\frac{t_{\text{naturalLU}}}{t_{\text{reorder}}+t_{\text{LU}}}
\label{eq:speedup}
\]

\subsection{Reasonablity of End-Max Chain Loss}
\begin{figure}[!htbp]
    \centering
    \subfloat[End-max Chain Loss on Training set]{
    \includegraphics[width=0.5\textwidth]{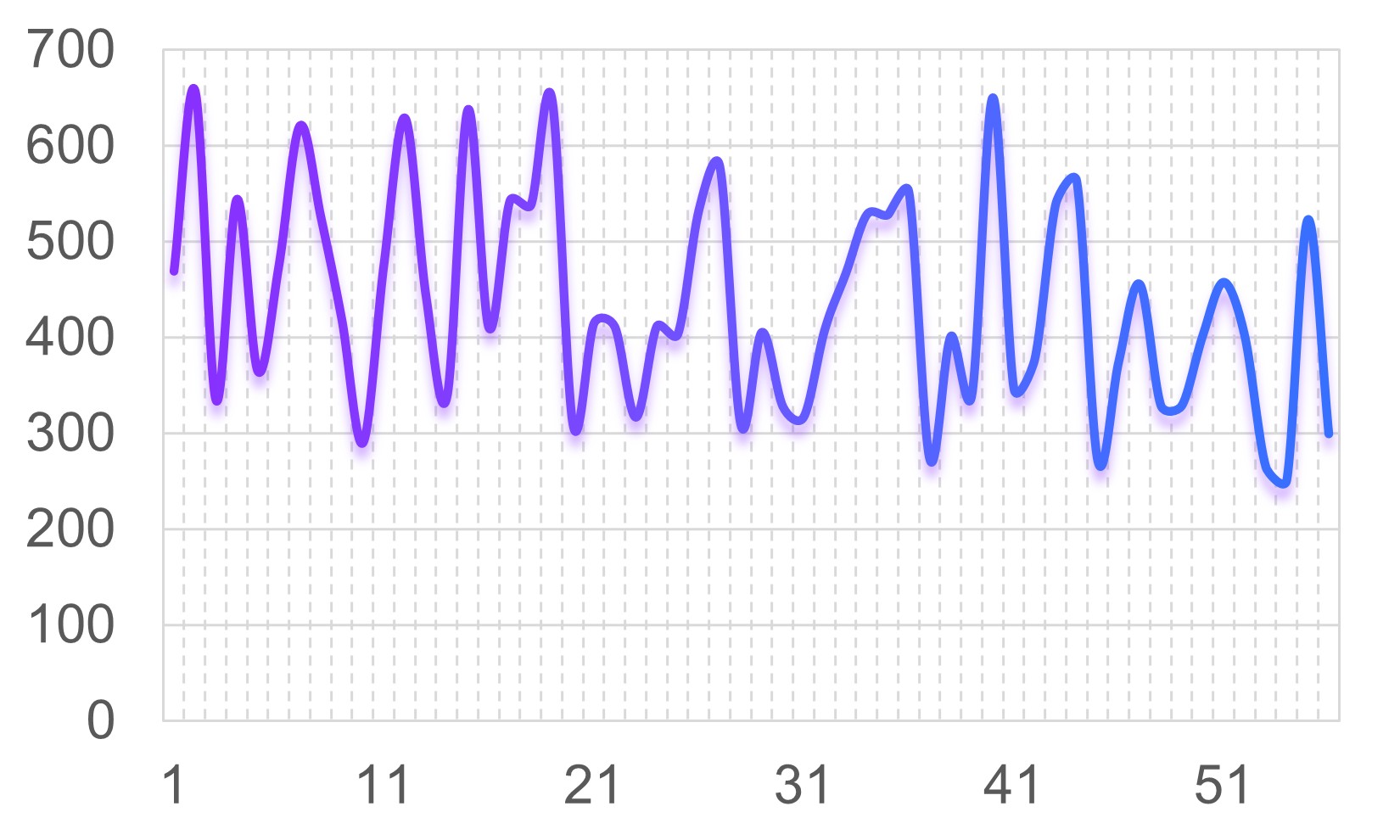}
    }
    \subfloat[Fill-in Ratio on Training set]{
    \includegraphics[width=0.5\textwidth]{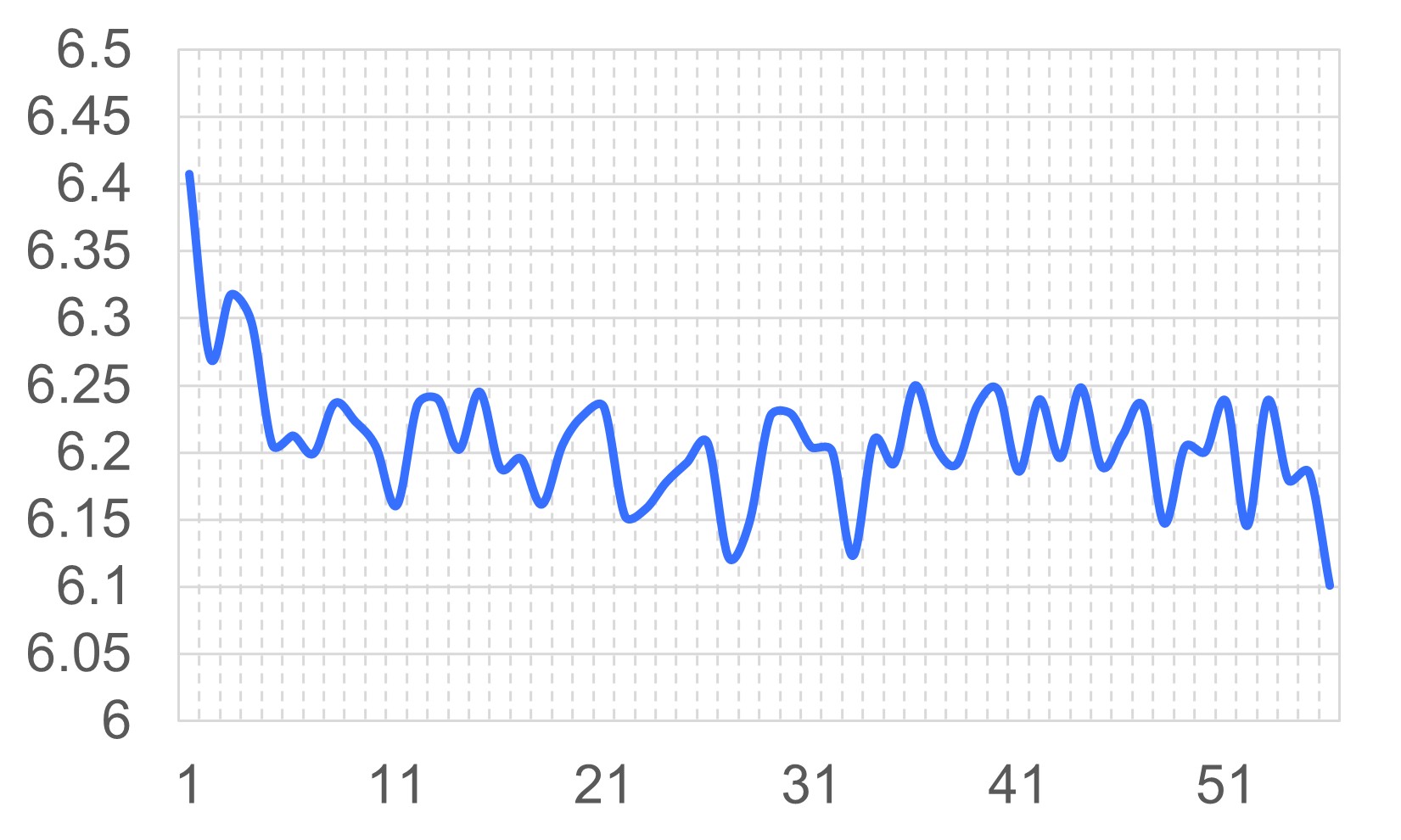}
    }
    \caption{Learning Curves of the Proposed Method with the MultigridSAGE backbone and learning rate \(10^{-5}\) on the Benchmark Training Set.}
    \label{fig:learncurve}
\end{figure}
Both the end-max chain loss and the fill-in ratio are reported per epoch on the training set shown in Fig.~\ref{fig:learncurve}. From a local perspective in Fig.~\ref{fig:learncurve}(a), the proposed loss curve goes up and down in a local range due to the large triplet sampling space. However, the global trend of the proposed loss decreases steadily along with the training epochs. This coincides with the fill-in ratio reduction curve on the training set. The similarity of their global declining trends indicates a certain reasonability of the proposed loss function.

\subsection{Fill-in and Speedup Analysis}

Reordering model learned from both theoretical and empirical guaranteed loss is evaluated on five randomly sampled matrix subsets from benchmark sparse matrix collection SuiteSparse. Each subset corresponds to a scientific problem category denoted CFD, MRP, SP, 2D/3D and TP. Each entry in both Table~\ref{tab:fillin_ratio} and \ref{tab:lu_time} is computed as the arithmetic mean over all the matrices in the corresponding set. The bold type value means the best and the underlined value means the best from baselines. For comprehensive and fair analysis, we compute the speedup metric using the sum of the LU factorization time and the reordering time.
\begin{table}[!htbp]
\centering
\begingroup
\footnotesize
\setlength{\tabcolsep}{8pt} 
\caption{Fill-in ratio for various ordering methods on the benchmark test set.}
\label{tab:fillin_ratio}
\begin{tabular}{|l|rrrrrr|}
\hline
& CFD & MRP & SP & 2D3D & TP  & All \\
\hline
Natural
&361.23 &83.46 &185.60 &474.71 &523.32  &263.19 \\
\hline
AMD
&386.75 &402.85 &194.35 &544.43 &642.35  &337.36 \\
Metis
&73.49 &59.37 &56.02 &85.62 &93.75  &66.16 \\
Fiedler
&\underline{\textbf{56.42}} &\underline{\textbf{46.28}} &48.39 &\underline{78.23} &\underline{69.59}  &\underline{54.58} \\
\hline
$S_e$
&61.62 &52.09 &55.66 &79.95 &72.41  &60.27 \\
GPCE
&61.61 &47.89 &49.98 &80.54 &70.33  &57.10 \\
UDNO
&59.57 &48.25 &\underline{46.67} &79.75 &75.41  &55.38 \\
CFP
&57.87 &47.76 &\textbf{45.63} &\textbf{74.85}&\textbf{52.84}   &\textbf{52.76} \\
\hline
\end{tabular}
\endgroup
\end{table}

\begin{table}[!htbp]
\centering
\begingroup
\footnotesize
\setlength{\tabcolsep}{8pt}
\caption{Speedup Analysis of Various Matrix Reordering Algorithms in LU Factorization of Benchmark Test Matrices. Natural means direct LU factorization time in seconds without reordering.}
\label{tab:lu_time}
\begin{tabular}{|l|rrrrrr|}
\hline
& CFD & MRP & SP & 2D3D & TP  & All \\
\hline
Natural &466.81&	72.81&	913.91&	734.79 &873.11& 649.32 \\
\hline

AMD &10.74& 1.01  & 1.63 &1.44 &0.73 &3.70 \\
Metis &46.67  &6.03 &	24.77 &	30.43 &	122.91 &	32.67\\
Fiedler &83.03 &	1.93 &	48.15 &84.67 &	62.63 &	54.46\\
\hline
$S_e$ & \underline{86.37} &6.51 &	49.82 &	89.63 &	402.16 & 73.94\\
GPCE &85.69 &	\underline{7.65} &	\underline{67.64} &	\underline{94.05} &	\underline{435.67} &	\underline{83.80}\\
UDNO &84.37 &	5.99 &	46.74 &	85.48 &	399.82 &	71.44 \\
CFP & \textbf{99.91} &	\textbf{7.83} &	\textbf{76.43} &	\textbf{91.90} &	\textbf{524.96} &	\textbf{95.23}\\

\hline
\end{tabular}
\endgroup
\end{table}
Fill-in ratios from matrices reordered by Fiedler scores in Fig.~\ref{fig:fillin}(a) are overwhelmingly less than other baselines including both graph-theoretic and deep learning methods. This indicates that spectral information is significant for reordering such matrices. Taking the approximate Fiedler score \(S_e\) as input, the proposed method reduces fill-ins further by encoding additional structural knowledge from the graph itself. Especially for SP and TP matrices, fill-in ratio of CFP is obviously lower than the state-of-the-art baselines. The comparison result shows its generalization capability because only 2D3D matrices are used for training, and none of the 2D/3D test matrices are seen during training. Moreover, CFP further reduces fill-ins compared with the best deep learning approach, UDNO, which shows the superiority of the proposed loss function. We conduct detailed analysis in the following subsection.

Fill-ins are used to pre-estimate the subsequent memory usage and computation. In real-world applications, the de facto running time is more important in Table~\ref{tab:lu_time}(b). In contrast to the fill-ins in Table~\ref{tab:fillin_ratio}(a), the deep learning baselines dominate because of their efficient reordering time. Even though their fill-ins are higher than some graph theoretic methods, their running times are promising to be more excellent.

\subsection{Ablation Study}
Multigrid GNNs and end-max loss function are two major components of CFP. To investigate the choice of basic block in this multigrid network, we replace graph sage unit SAGEConv in CFP with Graph Attention unit GATConv, i.e., the second row in Table~\ref{tab:ablation}. Keeping its backbone, the loss function of CFP is replaced with pairwise cross entropy under the supervision of the best ordering derived from AMD, Metis and Fiedler. Fill-ins of its predicted ordering on SP, CFP and SP+CFP are shown in the third row of Table~\ref{tab:ablation}. Similarly, CFP with UDNO loss function is in the fourth row. 
\begin{table}[!htbp]
\caption{Ablation Study of the Proposed Method CFP.}\label{tab:ablationloss}
\centering

\footnotesize

\begin{tabular}{|c|c|c|c|ccc|}
\hline
 &input&network &loss function  & SP& CFD& SP+CFD\\
   \cline{2-4}
 \hline
 
1&\multicolumn{3}{|c|}{$S_e$} & 	55.66 &61.62&58.64\\
\hline
2&\(S_e\)&MgGAT&end-max chain&46.57&69.20&54.77\\
\hline
3&\(S_e\)&MgSAGE&PCE&51.48 	 	&59.80  &	54.46   \\
4&\(S_e\)&MgSAGE&UDNO&46.67 &	59.57 &	53.12 \\
\hline
&\(S_e\)&MgSAGE&end-max chain&45.63&57.87&50.07\\ 	 

\hline
\end{tabular}
\label{tab:ablation}
\end{table}

The large fill-in distinction between CFP and its adaption versions suggests the key role of the proposed loss function.
Similar to UDNO, CFP is a self-supervised learning framework, only relying on the sparsity structure of the input matrix. CFP derives a theoretical guaranteed loss function for optimization, while UDNO designs a loss function to make the reordered matrix satisfying intuitive property, such as make nonzeros concentrate on the diagonals. Compared with UDNO, fill-in reduction of CFP shows the proposed end-max chain loss is more elegant self-supervised learning framework without much prior knowledge.
\section{Conclusion}
In this work, we introduced CFP, a self-supervised sparse matrix reordering method that is consistent with the Fill-Path Theorem. By using the graph structure of the matrix and learning an elimination order with multigrid GNNs and an end-max chain loss, CFP aims to reduce fill-in at its source. In our experiments, CFP produced lower fill-in ratios and faster LU factorization times than representative graph-theoretic and learning-based baselines.

\subsubsection{\ackname}
This research was supported by the National Key R\&D Program of China under Grant No. 2021YFB0300203.

\end{document}